\documentclass[mlmain,twocolumn]{jmlr}

\usepackage{longtable}

\usepackage{booktabs}
\usepackage[load-configurations=version-1]{siunitx} 

\theorembodyfont{\upshape}
\theoremheaderfont{\scshape}
\theorempostheader{:}
\theoremsep{\newline}

\jmlrvolume{}
\firstpageno{1}

\title[Self-Supervised Pre-training via Time Reversal]{Self-Supervised Mental Disorder Classifiers via Time Reversal}

   \author{\Name{Zafar Iqbal} \Email{ziqbal@student.gsu.edu}\\
   \Name{Usman Mahmood} \Email{umahmood1@gsu.edu}\\
   \Name{Zening Fu} \Email{zfu@gsu.edu}\\
   \Name{Sergey plis} \Email{splis@gsu.edu}\\
   \addr Georgia State University, TReNDS}

\begin{document}

\maketitle{}

\begin{abstract}
Data scarcity is a notable problem, especially in the medical domain, due to patient data laws. Therefore, efficient Pre-Training techniques could help in combating this problem. In this paper, we demonstrate that a model trained on the time direction of functional neuro-imaging data could help in any downstream task, for example, classifying diseases from healthy controls in fMRI data. We train a Deep Neural Network on Independent components derived from fMRI data using Independent component analysis (ICA) technique. It learns time direction in the ICA-based data. This pre-trained model is further trained to classify brain disorders in different datasets. Through various experiments, we show that learning time direction helps a model learn some causal relation in fMRI data that helps in faster convergence, and the model generalizes well in downstream classification tasks even with fewer data records.

\end{abstract}

\section{Introduction}
\label{sec:intro}

The proliferation of Medical imaging modalities such as Computed Tomography (CT) \citep{garvey2002computed}, Positron Emission Tomography (PET) \citep{sudarshan2020joint} and Magnetic Resonance Imaging (MRI) \citep{ammar2021automatic} has made it possible to acquire large amounts of health informatics data. However, due to practical reasons, the data could not be shared publicly in most cases. Hence researchers are constrained to work with small datasets. Functional Magnetic Resonance imaging (fMRI) is a useful imaging technique widely used in investigating psychiatric behaviours and neuro-cognitive diagnosis \citep{yang2020current}. The fMRI is broadly classified into two parts; resting state fMRI and task-based fMRI. In the former, the subject is in a task negative state and in latter, the subject is asked to perform some cognitive task. Generally, fMRI is performed using a method called Echo Planar Imaging (EPI)\citep{mansfield1977multi}. It collects data for 2D image in approximately every 60 ms at a resolution of 3.4 x 3.4 x 4 $mm^3$ voxel size.To cover the whole brain, approximately 32 slices are acquired with repetition time of 2s/volume. \citep{glover2011overview}.

The high dimensional nature of the fMRI data makes it challenging to effectively train a deep learning model on it owing to high computational cost. Likewise, such high dimensional data is inherently inefficient due to redundant information (noise) in it. Therefore, reducing the dimensions of the data without losing useful information could be helpful during the 
training phase of the model. Functional brain networks derived from fMRI data can serve this purpose and those can be used as a potential biomarkers for mental disorders \citep{salman2019group}. One such technique to extract functional brain networks is called Independent component anlaysis (ICA) which is explained in section \ref{sec: lr}.

Our goal is to use ICA based data to pre-train a model from scratch in a self-supervised way on time direction and investigate its efficacy in the downstream classification task of three diseases, namely, Schizophrenia (SZ), Autism and Alzheimer's Disease (AD) from their respective healthy subjects. \figureref{fig:tdir} shows a simplistic example where the process of breaking a jar is demonstrated with respect to time. The Jar is in different states at different time points. The direction of time gives us information about the series of events happened. In a similar fashion, fMRI data is collected and processed. Typically, to acquire fMRI data, we get a series of 2D axial slices to cover the entire brain, which is called a volume. This process of acquiring volumes is repeated over time which enables us to look at the functions of the brain. We have showed that learning time direction (forward and backward sequence of volumes) can help in two ways: the model can generalize well even with fewer number of subjects for training and the training time in the downstream task decreases considerably. We demonstrate that a model trained on forward and reverse time direction can outperform the model trained from scratch.

\section{Literature Review}
\label{sec: lr}

Data scarcity is a well known problem in domains where data collection is an expensive and difficult process. One such domain is Medical imaging data\citep{izonin2021grnn}. There are many protocols that needed to be followed before making such data public. Owing to such restrictions, it is very challenging to get one's hands on enough medical data to train a model effectively\citep{kadam2020performance}. One of the methods used to help resolve this issue is unsupervised pre-training of the machine learning model. It acts as a regularizer and could help in better generalization in any downstream task \citep{erhan2010does}. Stacked denoising autoencoders (SDAE) \citep{vincent2010stacked} and Deep Beliefs Nets \citep{hinton2009deep} are two of many classical models. These methods, however, are not very popular outside the field of Natural Language Processing (NLP) \citep{goodfellow2016deep}. In Computer Vision (CV), supervised pre-training is widely used on large imaging datasets such as COCO or Imagenet. With medical imaging, however, We do not have such large datasets to pre-train a model. Therefore, self-supervised learning is a suitable method in such cases. It has been shown that self supervised methods outperform supervised methods when small datasets are used for pretraining \citep{deng2009imagenet}.

\begin{figure}[htbp]
\floatconts
  {fig:tdir}
  {\caption{Forward and reverse time direction of a broken Jar.}}
  {\includegraphics[width=1\linewidth]{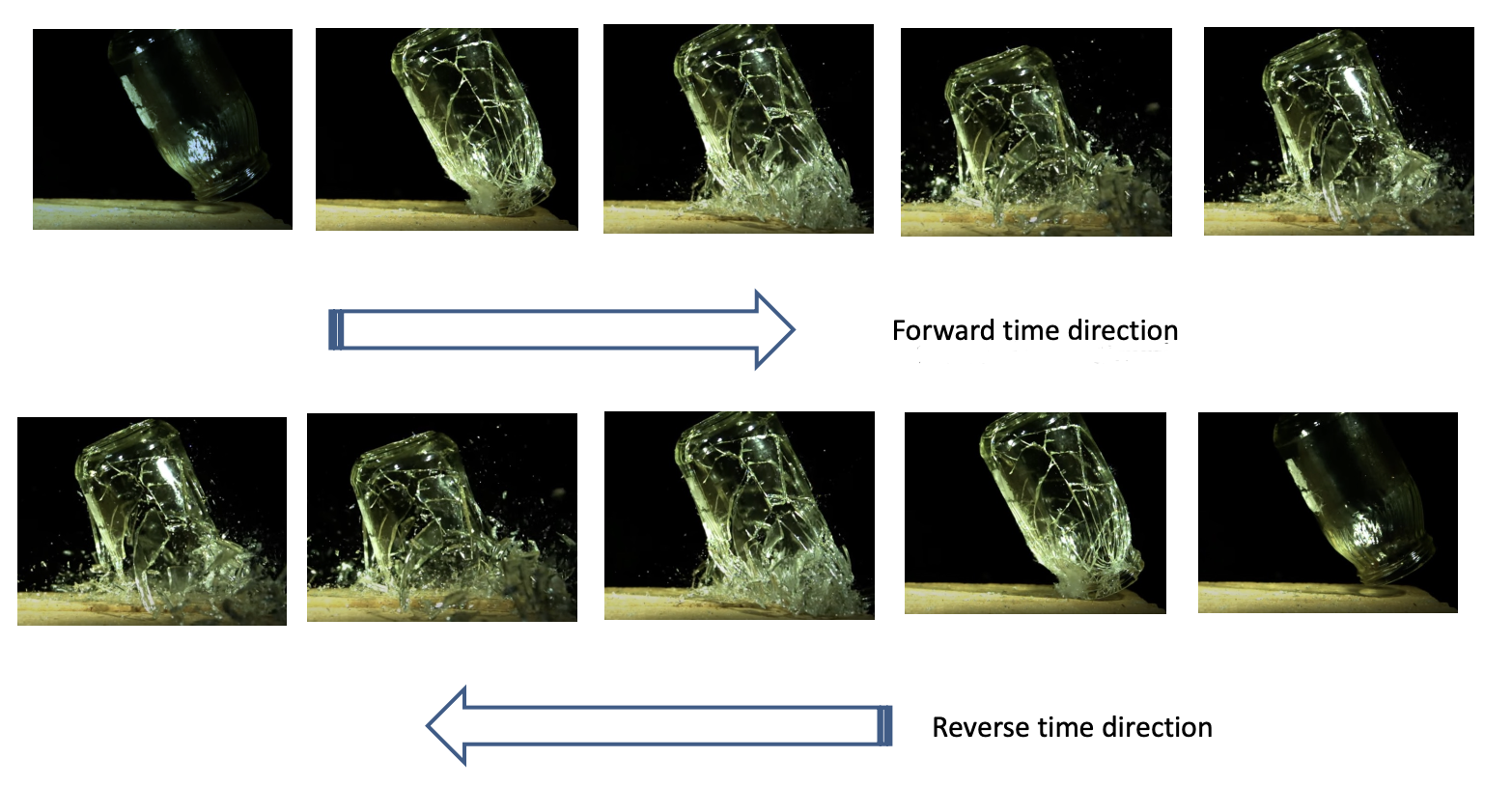}}
\end{figure}

Owing to the advancement of medical data acquisition techniques, it is possible to acquire high dimensional data \citep{faisal2021imputation}. Such data encompass a large amount of information about the subject. However, processing high dimensional data is computationally expensive and also inefficient due to redundancy and noise in the data \citep{wang2021advances}. Therefore, the dimensionality reduction techniques could be useful. One of the most widely used multivariate techniques to estimate brain functional networks from fMRI data is called independent component analysis (ICA). Unlike other techniques based on general linear model (GLM) \citep{moeller2011independent}, ICA does not depend on prior information in calculating time points. Furthermore, it could help de-noise the fMRI data by decomposing artifacts into independent components, shedding off noise and redundant information in the volumetric data \citep{salman2019group}.

\section{Methodology}
\label{sec:archi}

The proposed method includes two phases; the first phase is pretraining of the model on (Human Connectome Project (HCP) time courses and the second phase is transfer learning, where the pretrained model is further trained on four different datasets, described in section \ref{sec:math}. A generic block diagram of the proposed method is shown in \figureref{fig:tscheme}. To evaluate the performance of the model, we used Area Under the ROC curve (AUC) metric.  The Receiver Operator Characteristic (ROC) is a probability curve that plots True Positive Rate (TPR) against False Positive rate (FPR) at different thresholds. The AUC score is therefore a measurement of the ability of a model to distinguish between two classes. The performance of the model is shown in section \ref{sec:datasets}.

\begin{figure}[htbp]
\floatconts
  {fig:tscheme}
  {\caption{Training Scheme.}}
  {\includegraphics[width=1\linewidth]{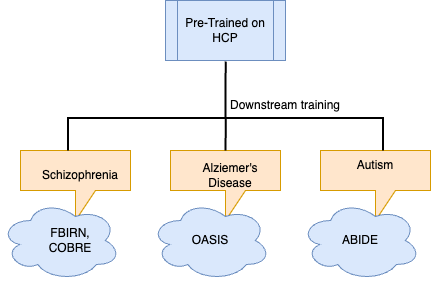}}
\end{figure}

We have used three 1D convolutional layers followed by a fully connected with 256 units and a Bi-directional long short-term memory (LSTM) layer and two more fully connected layers. We feed ICA time courses to the convolutional layers in the form of a sequence of windows. The output features used for the convolutional layers are 64, 128 and 200 respectively and the kernel sizes used in  the first two layers are 4 each and in the third convolutional layer is 3. Leaky Rectified Linear Unit (Relu) is used as an activation function in all the layers except in the last layer where we used Softmax function to convert the scores to a normalized probability distribution. The convolutional layers essentially act as an encoder. The encoder encodes the input data into latent representations z. The latent representation of the entire time series is then fed to a Bi-directional LSTM layer with 200 hidden units in sequence. The output from biLSTM layer is fed to attention layer to get the representation of entire time series in the form of a single vector c. The output is then passed through two fully connected layers to get the classification scores as shown in \figureref{fig:architecture}. Further details of the architecture can be found from this research article \citep{mahmood2020whole}.

\begin{figure}[htbp]
\floatconts
  {fig:architecture}
  {\caption{Architecture of the model.}}
  {\includegraphics[width=1\linewidth]{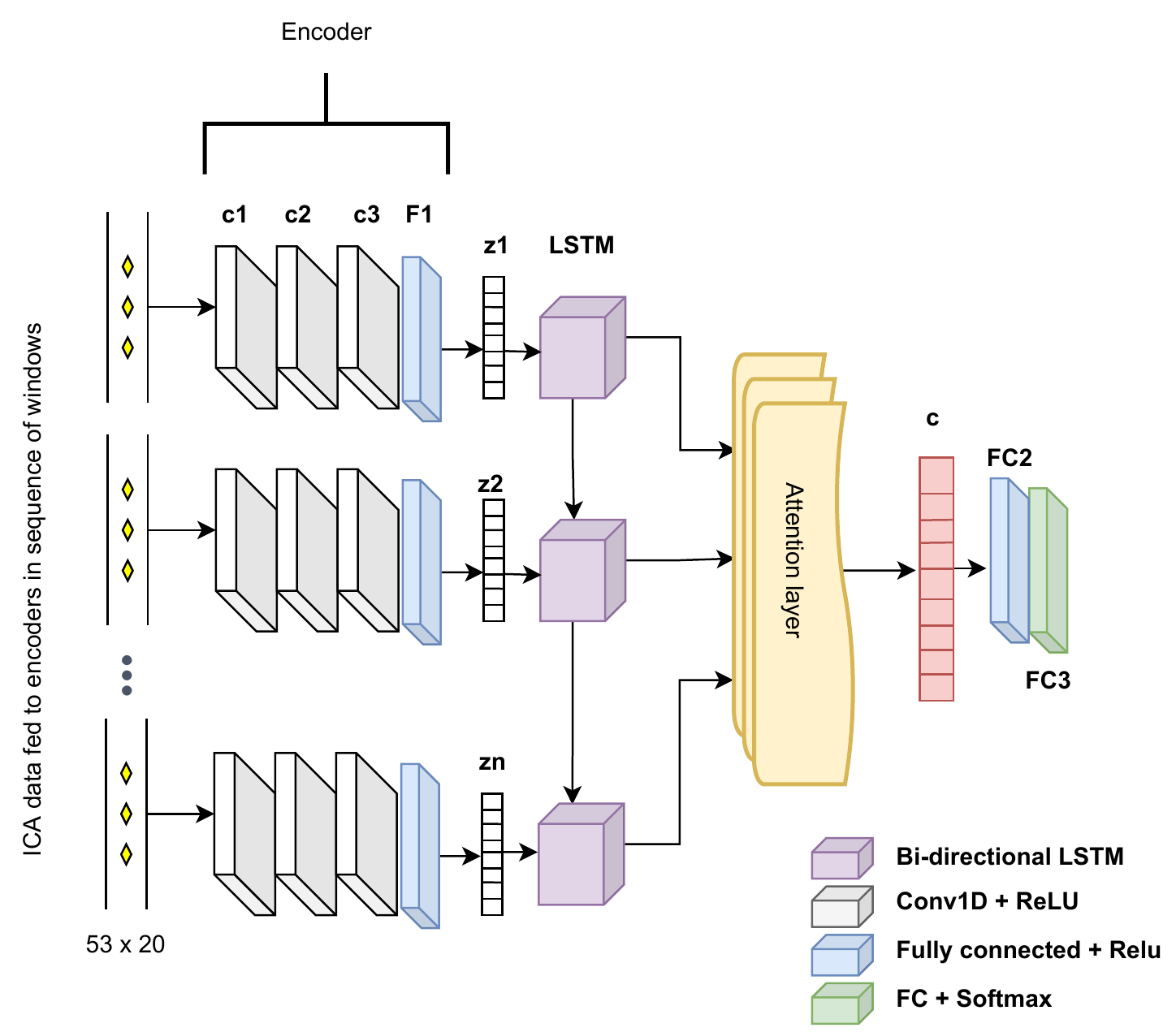}}
\end{figure}

\section{Experiments and Results}
\label{sec:math}

In this section we have described the performance of the Pre-Trained (PTR) model on the datasets described in section \ref{sec:datasets}. The model is pretrained on time direction of subjects from Human connectome Project (HCP) \citep{hcp}. The pretrained model is then further trained on four datasets for the downstream classification task of three abnormalities, namely, Autism, Schizophrenia and Alzheimer's diseases from healthy controls in the respective datasets. Schizophrenia is a severe psychotic mental disorder. The symptoms include diminished emotional expressions, the lack of motivation, paucity of speech etc. Alzheimer's Disease is also a neurological disorder which is a progressive form of Dementia. It leads to declining of memory functions and deteriorating social and behavioural functions. Autism leads to severe social-emotional reciprocity, affects non-verbal communicative behaviors and also impacts the ability to understand and maintain relationships. Interestingly, all these three mental disorders have common clinical features, as they present some kind of impairment in the cognitive ability \citep{levine2020clinical}. 

For both pre-training and the downstream task of classification, we used the same architecture as shown and described in section \ref{sec:archi}. The datasets used for downstream tasks are collected from the projects of Functional Biomedical Informatics Research Network (FBIRN) \citep{fbirn}, Centre of Biomedical Research Excellence (COBRE) \citep{cobre}, Autism Brain Imaging Data Exchange(ABIDE) \citep{abide} and Open Access Series of Imaging Studies (OASIS) \citep{oasis}. To evaluate how well the pretrained model performed, we also trained a model from scratch for the downstream classification task. The results of the experiments conducted are discussed in section \ref{sec:disc}. 

\subsection{Datasets}
\label{sec:datasets}
For pretraining, we used the ICA based data from HCP project. It has 823 subjects in total. The pretraining was done based on the time direction in FMRI data. We labelled the normal time direction as 0 and then reversed the time direction in each component and labelled it as 1. By adding up both the classes, we got 1646 subjects with 823 representing forward time direction and the other 823 subjects representing reverse time direction. The model was able to learn order of time with high precision.

We used four datasets to evaluate the performance of the PTR model on the classification task. For Schizophrenia classification, we used COBRE and FBIRN. For Autism we trained the model on ABIDE and for the classification of subjects with Alzheimer's disease, we used OASIS. It is pertinent to mention here that all the datasets used in this study are Pre-Processed through a method called Independent Component Analysis (ICA). The fMRI data is processed using Statistical Parametric Mapping Working (SMP12). Subsequently, only subjects with head motion of  $\leq 3^{\circ} $ and $\geq 3 mm $ were chosen \citep{fu2019altered}. For all the datasets under study, we acquired 100 ICA components and only 53 non-noise components were used for the training purposes. ICA based data is handy because of its lower dimensional nature. Unlike techniques such as General linear model, this approach requires no prior information in marking the regions. One important advantage of ICA based data is that it contains less noise then the actual FMRI data which helps getting good performance from the model \citep{gica}.
\subsubsection{Schizophrenia}
\label{SZ}
For the classification of Schizophrenia (SZ), two datasets were used; FBIRN and COBRE.

In FBIRN, There are 311 subjects in total. The number of subjects possess SZ tallies to 160 and the remaining 151 are the healthy controls (HC). Each subject has 53 non-noise components with 140 time points in each component. For experimentation, we used non-overlapping windows of size 53 X 20. The number of windows calculated to be 7 given that there were 140 time points. To evaluate the performance of the model on the said dataset, k-fold cross validation is applied. The validation and test data size was 59 each.

\begin{figure}[htbp]
\floatconts
  {fig:fbirn}
  {\caption{AUC Scores for PTR and Not-pretrained (NPT) models trained on FBIRN.}}
  {\includegraphics[width=1\linewidth]{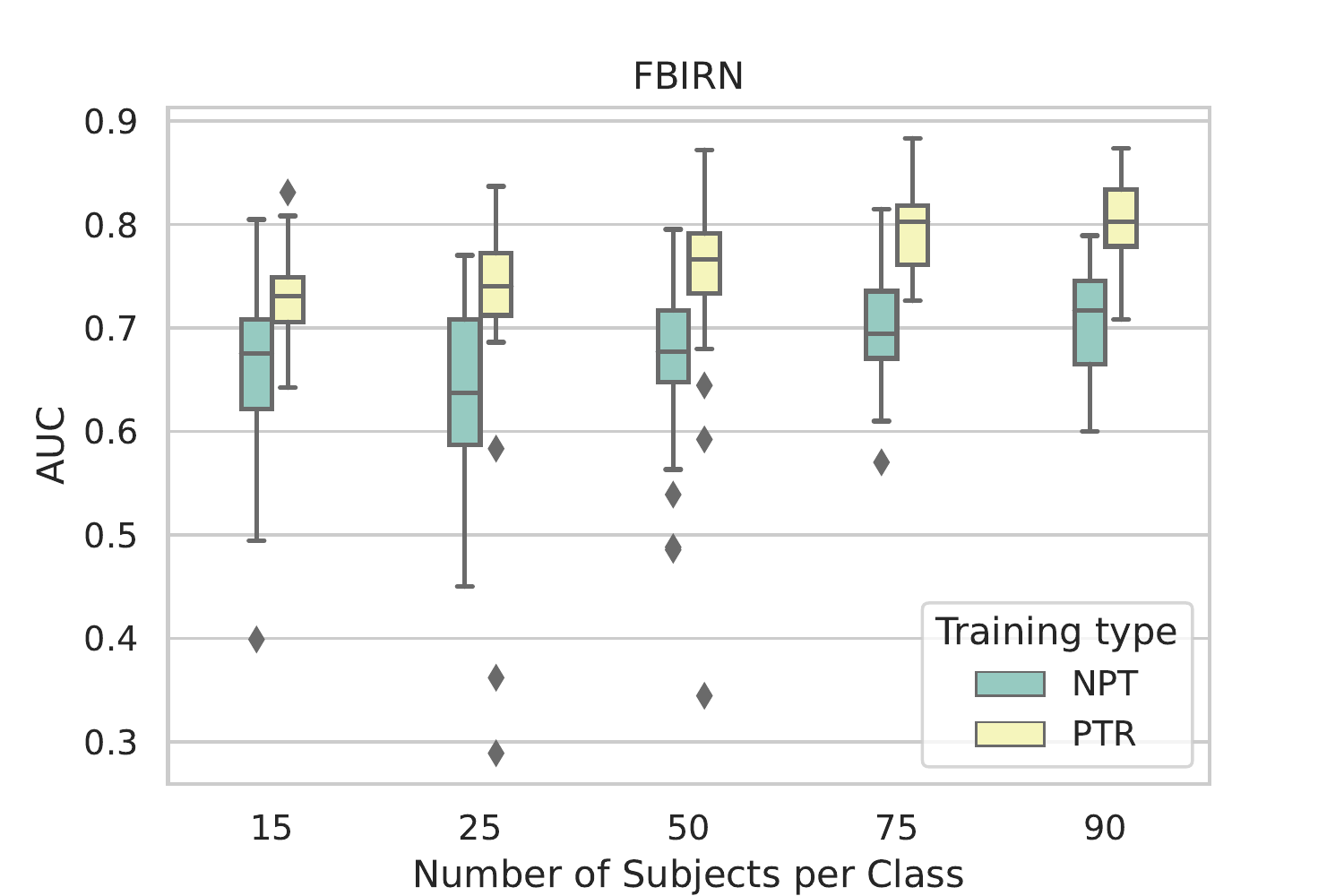}}
\end{figure}

Similarly, COBRE has 157 subjects in total. The number of subjects affected by SZ totals to 89 and the remaining 68 are the healthy controls (HC). Similar to FBIRN, Each subject has 53 non-noise components with 140 time points in each component. For experimentation, we used non-overlapping windows of size 53 X 20. The number of windows calculated to be 7 given that there were 140 time points. To evaluate the performance of the model on the said dataset, k-fold cross validation is applied. Two holdout datasets of size 27 each were used for validation and testing of the trained model.

\begin{figure}[htbp]
\floatconts
  {fig:cobre}
  {\caption{AUC Scores for PTR and NPT models trained on COBRE.}}
  {\includegraphics[width=1\linewidth]{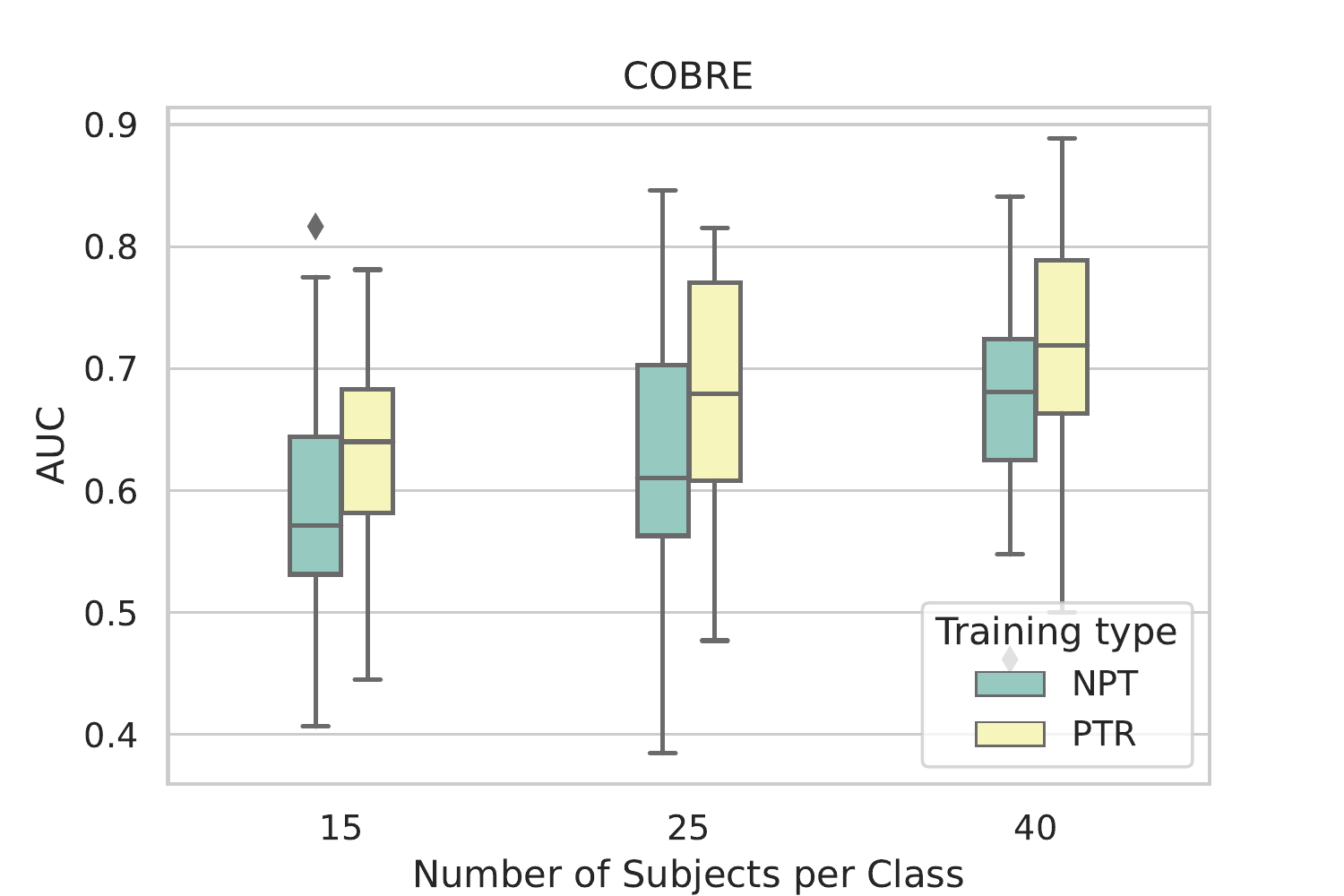}}
\end{figure}

\subsubsection{Alzheimer's Disease}
\label{AZ}
For the classification of Alzheimer's Disease, we used OASIS dataset.It has a total of 823 subjects, out of which 651 are Healthy Controls and the remaining 172 are affected with Alzheimer's disease. Like FBIRN or COBRE, Each subject has 53 non-noise components with 140 time points in each component. For experimentation, we used non-overlapping windows of size 53 X 20. The number of windows calculated to be 7 with 140 time points. The dataset is imbalanced given that the HC class has way more records then the other one. To tackle the class imbalancing issue, we ran multiple trials by using all the subjects with Alzheimer's disease and 172 subjects from HC class in sequence to make sure all the subjects are used in training and evaluating the model
To evaluate the performance of the model on the said dataset, k-fold cross validation is applied. Two holdout datasets of size 69 each were used for validation and testing purposes.

\begin{figure}[htbp]
\floatconts
  {fig:oasis}
  {\caption{AUC Scores for PTR and NPT models trained on OASIS.}}
  {\includegraphics[width=1\linewidth]{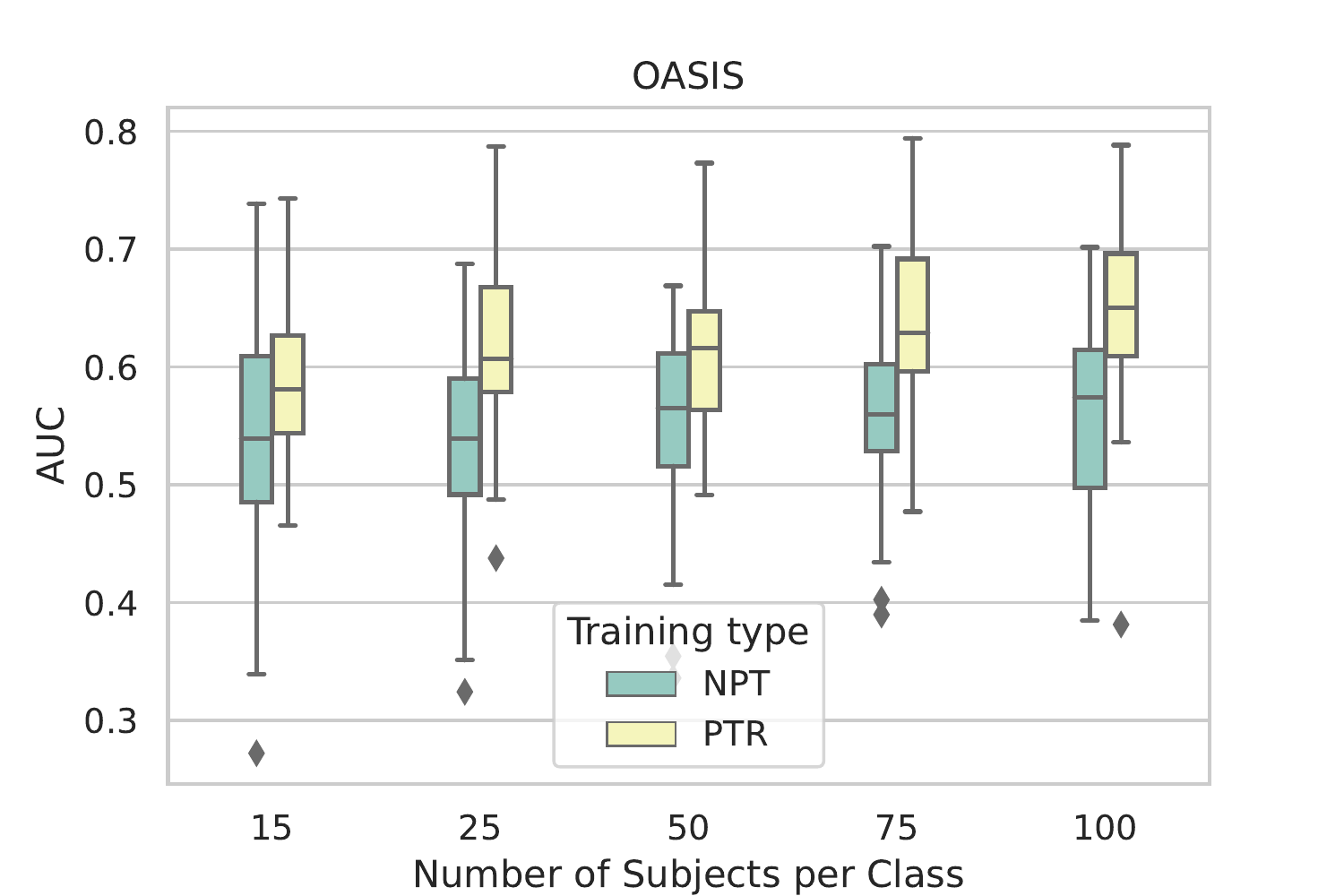}}
\end{figure}

\subsubsection{Autism}
\label{AU}
The dataset ABIDE has 869 subjects.The number of subjects affected by Autism totals to 471 and the remaining 398 are the healthy controls (HC). Similar to other datasets, Each subject has 53 non-noise components with 140 time points in each component. For experimentation, we used non-overlapping windows of size 53 X 20. The number of windows calculated to be 7 given that there were 140 time points. To evaluate the performance of the model on the said dataset, k-fold cross validation is applied. Two holdout datasets of size 237 each were used for validation and testing purposes.

\begin{figure}[htbp]
\floatconts
  {fig:abide}
  {\caption{AUC Scores for PTR and NPT models trained on ABIDE.}}
  {\includegraphics[width=1\linewidth]{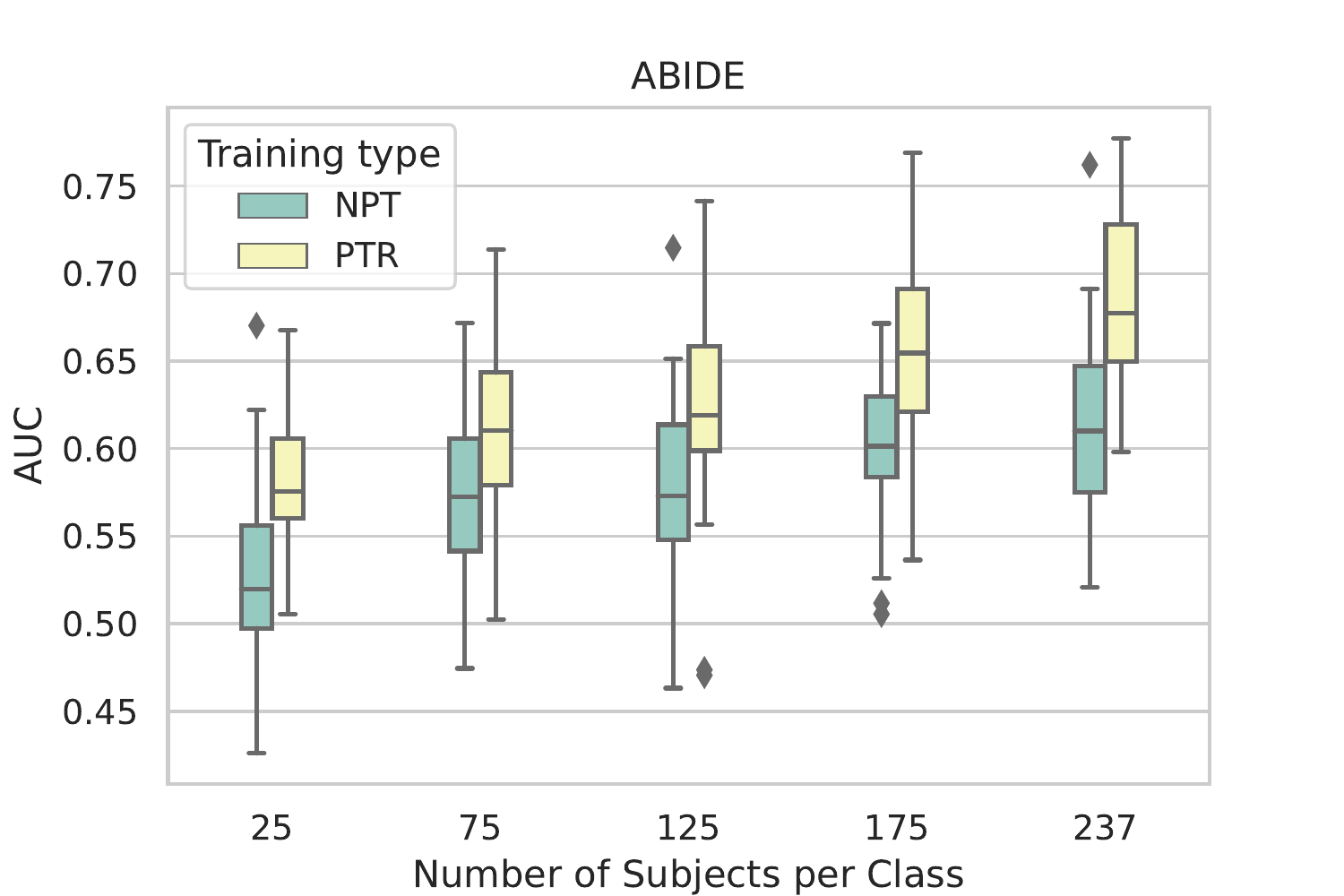}}
\end{figure}

\section{Results and Discussion}
\label{sec:disc}

Our hypothesis was that training a model on time direction of fMRI data could help in learning hidden dynamics of the data which in turn could help in any downstream task. To prove our hypothesis, we pretrained the model on HCP time courses. We had 53 non-noise components in each subject in the HCP dataset.We reversed the time points in each component and pretrain the model on forward and reverse time direction. We evaluated the classification ability of the pretrained model using four datasets. We experimented with different number of subjects per class to observe how well the model performs even with fewer number of subjects to train.

For Schizophrenia classification, We used two datasets (FBIRN and COBRE) to further train the model. \figureref{fig:fbirn} shows the details of classification result on FBIRN. It is noticeable that even with fewer number of subjects during training phase, the pretrained model generalizes well and the difference between the AUC scores obtained on test data in both the setups is reasonably large. When we increase the number of subjects, the pretrained model gets more data for training and thus the performance also increases significantly. We see an increase in the mean AUC score as evident from the \figureref{fig:fbirn}. The mean and median AUC score with pretraining comes as 0.801 and 0.802 respectively, while without pretraining the respective values are  merely 0.70 and 0.71.   For COBRE, We used the same setup. The results adhere with the previous findings, that is, the pretrained model was able to outperform the NPT model in SZ classification task, as shown in \figureref{fig:cobre}. With increasing number of subjects per class, we see a proportional increase in the AUC scores.

For Autism Vs Healthy Controls classification, we used ABIDE. In line with the previous results, the pretrained model is able to classify better even with 25 subjects in comparison to the model trained from scratch. As the number of subjects increase, the model starts performing well. An important point to notice here is that even with 237 subjects available for training, the NPT model has a median AUC score $\approx$ .61 in comparison to 0.67 when pretrained model was used. The age ranges in HCP and ABIDE are quite different with means 29.2 and 17.04 respectively \citep{mahmood2019learnt}, which proves that the pretrained model learnt signal dynamics that helped in the downstream task even with visible differences in both the datasets. Please refer to \figureref{fig:abide} for the results.

The dataset OASIS is used for classification of Alzheimer's Disease vs Healthy Controls. As shown in \figureref{fig:oasis}, there is not a visible difference when we choose 15 subjects for training. The reason behind this, in our understanding, is the differences in both the datasets. However, with increasing number of subjects per class, the model's performance gets better and the difference in AUC score between PTR and NPT models increase significantly.

\section{Conclusion}
\label{sec:conc}
 
 In this paper, we have investigated the ability of the pretrained model to classify abnormalities from health controls. We have demonstrated that a self supervised pretrained model on the time direction of fMRI data can learn information that could help classify better. We pretrained the model on HCP time points. The time direction was reversed and used both forward and reversed time directions for pretraining. The performance of the pretrained model is evaluated using the datasets described in sections \ref{AU}, \ref{AZ} and \ref{AU}. We have demonstrated that training on time direction of fMRI data using ICA time courses, in a self supervised manner, gives significant improvement in the downstream classification task. Pre-training with time reversal provides benefits that transfer across datasets. Learning dynamics of fMRI helps to improve performance of the model, as observed from the results discussed earlier.The Pre-trained model outperforms the NPT model significantly even with fewer number of subjects used for training. In future, we intend to skip ICA preprocessing and work directly with fMRI data.

\bibliography{references}

\end{document}